\documentclass[pdflatex,sn-mathphys-num]{sn-jnl}

\usepackage{graphicx}%
\usepackage{multirow}%
\usepackage{amsmath,amssymb,amsfonts}%
\usepackage{amsthm}%
\usepackage{mathrsfs}%
\usepackage[title]{appendix}%
\usepackage{xcolor}%
\usepackage{textcomp}%
\usepackage{manyfoot}%
\usepackage{booktabs}%
\usepackage{algorithm}%
\usepackage{algorithmicx}%
\usepackage{algpseudocode}%
\usepackage{listings}%
\usepackage{graphicx}
\usepackage{tikz}
\usepackage{caption}
\usepackage{subcaption}
\usepackage{booktabs}
\usepackage{colortbl}
\usepackage{float}

\usepackage{newtxtext,newtxmath}
\usepackage[T1]{fontenc}

\theoremstyle{thmstyleone}%
\theoremstyle{thmstyletwo}%

\theoremstyle{thmstylethree}%

\begin{document}

\title[Recognition of Daily Activities through Multi-Modal Deep Learning for AAL]{Recognition of Daily Activities through Multi-Modal Deep Learning: A Video, Pose, and Object-Aware Approach for Ambient Assisted Living}


\author*[1]{\fnm{Kooshan} \sur{Hashemifard}}\email{k.hashemifard@ua.es}

\author[1]{\fnm{Pau} \sur{Climent-Pérez}}\email{pau.climent@ua.es}

\author[1,2,3]{\fnm{Francisco} \sur{Florez-Revuelta}}\email{francisco.florez@ua.es}

\affil[1]{\orgdiv{Research Group on Ambient Intelligence for Active and Healthy Ageing, Department of Computing Technology}, \orgname{University of Alicante}, \orgaddress{\street{Ctra. San Vicente del Raspeig, s/n}, \postcode{03690},  \city{San Vicente del Raspeig}, \country{Spain}}}
\affil[2]{\orgname{Alicante Institute for Health and Biomedical Research (ISABIAL)}, \orgaddress{\street{Avda. Pintor Baeza, 12}, \postcode{03010}, \city{Alicante}, \country{Spain}}}\affil[3]{\orgname{ValgrAI - Valencian Graduate School and Research Network of Artificial Intelligence}, \orgaddress{\street{Camí de Vera s/n},  \postcode{46022}, \city{Valencia}, \country{Spain}}}


\abstract{Recognition of daily activities is a critical element for effective Ambient Assisted Living (AAL) systems, particularly to monitor the well-being and support the independence of older adults in indoor environments. However, developing robust activity recognition systems faces significant challenges, including intra-class variability, inter-class similarity, environmental variability, camera perspectives, and scene complexity. This paper presents a multi-modal approach for the recognition of activities of daily living tailored for older adults within AAL settings. The proposed system integrates visual information processed by a 3D Convolutional Neural Network (CNN) with 3D human pose data analyzed by a Graph Convolutional Network. Contextual information, derived from an object detection module, is fused with the 3D CNN features using a cross-attention mechanism to enhance recognition accuracy. This method is evaluated using the Toyota SmartHome dataset, which consists of real-world indoor activities. The results indicate that the proposed system achieves competitive classification accuracy for a range of daily activities, highlighting its potential as an essential component for advanced AAL monitoring solutions. This advancement supports the broader goal of developing intelligent systems that promote safety and autonomy among older adults.}

\keywords{Human Activity Recognition, Activities of Daily Living, Ambient Assisted Living, Multi-modal Fusion}



\maketitle

\section{Introduction}\label{sec1}

The demographic shift towards an aging global population presents an increasing need for technologies that support older adults to live independently and safely within their own homes \citep{WHO2024Ageing}. Ambient Assisted Living (AAL) systems aim to address this challenge by providing solutions that monitor well-being, improve safety, and offer timely assistance when needed. Automatic recognition of daily activities is therefore crucial for developing intelligent monitoring systems that can adapt their behavior based on situational needs, providing detailed information when necessary for safety while maintaining privacy during routine daily activities. 
In this domain, indoor Human Activity Recognition (HAR) specifically addresses activities carried out in confined environments, making it particularly relevant for AAL applications that involve continuous monitoring.

This work focuses on the recognition of activities of daily living (ADL) for older adults, which varies across diverse indoor environments such as homes, hospitals, care facilities, and assisted living centers. These activities range from basic ones like walking, sitting, sleeping, and drinking water, to complex multi-step sequences such as "taking medication" which can involve a series of simpler activities like "opening the pill box" and "drinking water," or "preparing meals" which may include food handling activities such as cooking and stirring~\citep{bhola2024review}. Our approach contributes to advancing such recognition capabilities specifically for context-aware privacy-preserving monitoring systems in residential settings.

Indoor activities present several key challenges: \textbf{intra-class variability}, which occurs when different individuals perform the same activity in distinct ways (for instance, "drinking water" while sitting, standing, or walking may generate different recognition patterns), \textbf{inter-class similarity} where distinct activities share comparable motion patterns (such as stirring tea versus stirring soup, or jumping and dancing which may have similar movements that are ambiguous for recognition systems), \textbf{view variance} which significantly impacts recognition performance when cameras are positioned at different angles or heights, and \textbf{object interaction complexity} where many daily living activities are inherently defined by how humans interact with household objects. In addition, environmental factors such as varying lighting conditions, occlusions, camera perspective, and scene complexity further complicate robust recognition in real-world indoor settings.

The unique nature of indoor activities means that conventional methods designed for general HAR may not be suitable, as the activity set differs from one environment to another, and one model cannot fit all scenarios. This specificity makes indoor HAR a distinct research track that requires activity recognition problems to be considered differently based on the application area, particularly for concepts such as smart homes, automated patient care, and older adults care systems.

Advances in computer vision and deep learning have provided powerful tools for HAR. 3D Convolutional Neural Networks (CNNs) have emerged as effective feature extraction backbones due to their ability to capture both spatial and temporal patterns in video data through hierarchical feature learning \citep{carreira2017quo}. However, traditional 3D CNNs with uniform spatio-temporal kernels applied across entire video volumes often lack the flexibility needed to address the nuanced challenges of indoor activities, particularly struggling with view variance where similar actions performed from different angles or with subtle variations are not well distinguished. To counteract this problem, human pose data has been used to provide view-invariant geometric information that remains consistent across different camera perspectives, making it valuable for robust activity recognition \citep{yang2021unik, duan2022dgstgcn}. Graph Convolutional Networks (GCNs) for analyzing skeletal pose information have shown promise in capturing spatio-temporal dynamics, particularly for addressing view variance challenges \citep{shi2019skeleton, yan2018spatial}. However, pose-based methods often struggle to differentiate activities with similar poses but different contexts, as they fail to effectively leverage complementary visual features and object interactions present in video data, limiting their effectiveness for fine-grained activity discrimination.

Multi-modal systems that integrate information from various sources such as video, pose, and object interactions are increasingly being explored to enhance contextual understanding and improve the disambiguation of activities \citep{das2021vpn++}. The integration of object context is particularly relevant for indoor daily activities, where the manipulation of objects often defines the action itself \citep{ryoo2020assemblenet++}, although it has received comparatively less attention in previous research.

This paper presents a multi-modal daily activity recognition system specifically designed for older adults in indoor AAL environments. The system leverages a 3D CNN to extract features from video streams, a GCN to model a dynamic human pose from 3D human pose (skeleton) data, and an object detection module to incorporate information about object interactions. Subsequently, these diverse data streams are fused using a cross-attention mechanism, enabling the model to focus on the most important features for activity classification. The proposed framework is evaluated on the Toyota SmartHome dataset~\citep{das2019toyota}, which features a range of unscripted daily activities performed by older individuals in a homelike setting, well suited to our AAL application domain.

The contributions of this work are threefold, each addressing specific challenges in indoor AAL activity recognition:

\textbf{First, we propose a multi-modal architecture that integrates visual information (3D CNN), human pose data (GCN), and scene context (object detection) through a cross-attention mechanism.} This directly addresses the challenge that many indoor activities are fundamentally defined by object interactions, as our system can distinguish between similar motions based on the objects involved, improving recognition of daily living activities where context matters.

\textbf{Second, we utilize a spatial embedding approach that establishes correspondences between 3D pose data and visual features.} This tackles the view variance problem common in indoor monitoring by aligning pose information (which is naturally view-invariant) with visual features, enabling our system to maintain recognition accuracy across different camera positions and angles typical in home environments.

\textbf{Third, our 3D CNN backbone with cross-attention enhancement learns both local motion patterns and global activity structure while being guided by object context.} This addresses the limitation of traditional CNNs that apply uniform processing across video volumes. Our approach adapts feature processing based on relevant objects and pose information, leading to more discriminative representations for complex indoor activities.

Consequently, this work advances accurate, context-aware activity recognition for AAL, a prerequisite for effective and privacy-sensitive monitoring that supports older adults in daily life. Continuous observation can threaten dignity and autonomy, so systems must adapt what they sense and store to the situation. Context-aware methods (e.g., \citep{padilla2015visual}) estimate salient variables such as appearance (e.g., degree of nudity), ongoing events (e.g., a fall), and current activity. HAR is paramount to this estimation: by inferring what a person is doing, the system can regulate the intensity of monitoring and the granularity of data retained. Our approach follows this principle, providing detailed analysis only when safety demands it and defaulting to lightweight, low-visibility operation during routine activities. In doing so, it balances safety with privacy, offering actionable assistance while minimising unnecessary data exposure.

The remainder of this paper is organized as follows. Section 2 will discuss related work in human activity recognition, particularly focusing on multi-modal approaches and systems applicable to AAL scenarios. Section 3 will detail the architecture of the proposed activity recognition system. Section 4 will describe the experimental setup, the dataset used, and the evaluation metrics. Section 5 presents and discusses the results, and finally, Section 6 concludes the article and suggests directions for future research.

\section{Related Work}\label{sec2}
Indoor HAR has advanced along four converging research threads: (i) strengthening spatio-temporal visual modeling beyond uniform 3D convolutions, (ii) leveraging skeleton streams to counter view variance while exposing motion geometry, (iii) incorporating object and scene context to disambiguate fine-grained ADL, and (iv) designing multi-modal fusion and cross-modal alignment mechanisms that go beyond late concatenation toward learned correspondence and attention. These lines are particularly important for ADL recognition, where inter-class similarity and intra-class variability meet strong object dependence and viewpoint changes~\citep{beddiar2020vision,bhola2024review}.

\subsection{Visual Backbone Evolution: From 3D CNNs to Transformers}

Classical 3D CNNs and multi-stream CNNs established strong baselines for video recognition, but their uniform spatio-temporal processing often blurs subtle discriminative cues and struggles with view shifts in cluttered indoor scenes~\citep{karpathy2014large, wang2015action, wang2016temporal}. Inflated 3D convolutions improved spatio-temporal modeling, yet still inherit sensitivity to camera geometry and weak grounding in object-centric semantics~\citep{carreira2017quo}. Non-local/self-attention operators add long-range dependencies but may diffuse focus in complex rooms when actor-relevant tokens are not explicitly guided~\citep{wang2018non, simonyan2014two}.

Recent video transformers such as ViViT address global context with space-time attention~\citep{bertasius2021space, arnab2021vivit}, but aside from the need for large datasets, uni-modal RGB remains prone to scene bias and viewpoint sensitivity in ADL. Alternative attention mechanisms have been explored, with Glimpse Clouds~\citep{baradel2018glimpse} using unstructured attention to extract visual glimpse sequences without spatial coherence constraints, leveraging pose supervision during training to focus attention on human anatomical structures.

\subsection{Skeleton-based Recognition: From Handcrafted Features to Graph Networks}

Skeleton-based human action recognition has evolved from handcrafted features to deep learning approaches. The early methods relied on depth sensors and spatiotemporal features for activity recognition~\citep{CHAARAOUI2014786, jalal2016human}. The introduction of Graph Convolutional Networks (GCNs) marked a significant advancement, with ST-GCN~\citep{yan2018spatial} first modeling skeleton data as spatiotemporal graphs with joints as vertices and natural body connections as edges.

Building upon ST-GCN, several improvements have been proposed. Directed Graph Neural Networks~\citep{shi2019skeleton} introduced directed acyclic graphs to better model kinematic dependencies between joints and bones. Two-Stream Adaptive Graph Convolutional Networks~\citep{shi2019two} combined first-order (joint coordinates) and second-order (bone vectors) information while learning adaptive graph topologies. Recent work has extended skeleton graphs with non-adjacent joint connections and improved partitioning strategies~\citep{wang2022skeleton}, while multi-scale approaches have been explored~\citep{lovanshi2024human}. Multi-stream architectures have also gained attention, with three-stream networks combining spatial-temporal graphs, tree-structure-reference-joints images, and RGB appearance data~\citep{fang20233}.

Graph self-attention networks like MGSAN~\citep{wang2024mgsan} demonstrate that combining GCNs with self-attention can capture intrinsic topology and long-term temporal dependencies between skeleton joints, moving beyond traditional 2D spatial plus 1D temporal approaches to true spatiotemporal modeling.

\subsection{Cross-modal Training Strategies: Skeleton-informed RGB Learning}

A central insight for indoor ADL is that skeletal information should shape the visual representation space during training, even when deployment relies solely on RGB input. Several approaches leverage pose information during training while maintaining RGB-only inference.

Pose-induced transformers~\citep{reilly2024just} explicitly inject 2D/3D skeleton priors into video transformers through auxiliary token–joint mapping and feature alignment, then remove pose modules at inference to retain efficiency. Built upon the TimeSformer backbone~\citep{bertasius2021space}, $\pi$-ViT introduces plug-in modules (2D-SIM and 3D-SIM) that perform pose-aware auxiliary tasks during training, enabling the video transformer to learn pose-augmented RGB representations without requiring poses during deployment. The approach implements cross-modal alignment as two auxiliary tasks: a token–joint mapping that teaches which visual tokens correspond to joints, and a skeleton feature alignment/classification that enforces RGB tokens to carry view-agnostic motion structure.

Complementary to this approach, self-supervised masked pretraining guided by skeleton latents likewise sculpts RGB representations to internalize motion regularities, enabling RGB-only fine-tuning and inference. For instance, SV-data2vec~\citep{dovzdor2025sv} employs a teacher-student architecture where the teacher generates contextualized targets from skeleton data while the student performs masked prediction on both skeleton and visual inputs, ultimately allowing fine-tuning with RGB data alone. After pretraining, the model fine-tunes with RGB-only, matching or surpassing RGB+pose methods on several protocols.

\subsection{Multi-modal Fusion and Attention Mechanisms}

Score averaging and feature concatenation are common fusion strategies, but they often fall short when RGB, pose, and object streams are misaligned in space, time, or geometry~\citep{yadav2021review, beddiar2020vision}. To tackle this problem, VPN and VPN++~\citep{das2020vpn, das2021vpn++} introduced enhanced video–pose integration modules that improve performance in indoor activity recognition.

Attention-based multi-stream architectures explicitly route information through learned connections to balance complementary modalities. Cross-modal transformers like STAR-transformer~\citep{ahn2023star} demonstrate the effectiveness of jointly processing video and skeleton features through specialized spatio-temporal cross-attention mechanisms, aggregating global grid tokens from video frames and joint map tokens from skeleton sequences into unified multi-class representations.

The Convolutional Block Attention Module (CBAM) has emerged as a particularly effective hybrid attention mechanism, combining channel and spatial attention to enhance feature extraction in multi-modal frameworks. CBAM-based approaches like CBAM-MFFAR~\citep{wang2024convolutional} achieve state-of-the-art performance by applying attention at multiple fusion stages, including early fusion through bidirectional lateral connections and late fusion of prediction scores.

Model-based fusion strategies have proven superior to simple concatenation or late fusion approaches. MMNet~\citep{bruce2022mmnet} demonstrates that transferring learned attention weights from skeleton modalities to guide RGB feature extraction significantly outperforms generic distillation methods, confirming that explicit correspondence between modalities matters more than simple logits transfer. Fusion-GCN extends this principle by directly incorporating additional sensor modalities into skeleton graph structures through either channel dimension fusion or spatial dimension fusion, achieving substantial improvements on large-scale multi-modal datasets~\citep{duhme2021fusion}.

Wearable sensor-based systems further demonstrate attention's versatility across modalities. Deep CNN-LSTM architectures with self-attention mechanisms achieve remarkable performance on smartphone sensor data, with models by learning temporal correlations between sensor readings~\citep{khatun2022deep}. TCN-attention frameworks leverage temporal convolutional networks with attention mechanisms to capture long-range dependencies in sensor time series, significantly outperforming traditional CNN-based approaches on multiple wearable sensor benchmarks~\citep{wei2024tcn}.

Contemporary 3D CNN approaches like spatiotemporal multi-modal learning frameworks demonstrate that attention-guided depth and pose representations can be effectively combined through two-stream architectures, where each modality provides complementary spatiotemporal information~\citep{wu2021spatiotemporal}.

\subsection{Object and Scene Context Integration}

Many ADL categories are fundamentally object-defined, so integrating detectors and modeling human–object relations is crucial~\citep{bhola2024review}. Region-based and one-stage detectors provide reliable cues that can be used to guide attention and token selection in video backbones~\citep{ren2015faster,redmon2016you}. This has been widely done in egocentric ADL recognition~\citep{grauman2022ego4d}; however, it is not as common in third-person vision-based HAR. AssembleNet++~\citep{ryoo2020assemblenet++} introduces peer-attention mechanisms that wire attention connections across modalities, with particular emphasis on object relations through omnipresent connectivity from object input blocks. Recent systems leverage pose-aware saliency to focus on hands and manipulated items; $\pi$-ViT's 2D token–joint supervision naturally complements object priors by tethering attention to anatomy-grounded regions.

Context-based representations can dominate over motion cues in action recognition, leading to scenarios where visually similar actions are distinguished primarily through background objects rather than human motion patterns~\citep{he2016human}. These training-time aligners are particularly effective against intra-class variability and inter-class similarity endemic to indoor ADL~\citep{bhola2024review}.

\section{Proposed Approach}\label{sec3}
Based on the literature review, indoor ADL benefits from: (i) strong spatio-temporal vision backbones, (ii) skeleton guidance (anatomy-aware) to stabilize against view variance and occlusions, (iii) object priors for human-object interactions, and (iv) cross-attention mechanisms that provide additional contextual information to mitigate intra-class variability and inter-class similarity. For elderly-focused indoor HAR, while transformers demonstrate strong performance in ADL, limited training data constrains their effectiveness. This motivates adopting cross-attention and correspondence mechanisms from transformers within more data-efficient architectures, leveraging their strengths while avoiding their data requirements.

Building on these insights, we propose a multi-modal activity recognition framework that integrates RGB video processed through a 3D CNN, 3D human pose data analyzed via a GCN, and object context information through pre-trained detection. These modalities are fused using cross-attention with explicit spatial embedding between pose and visual features to enable robust activity classification suitable for data-constrained elderly ADL scenarios.

\begin{figure}[t]
    \centering
    \includegraphics[width=\textwidth]{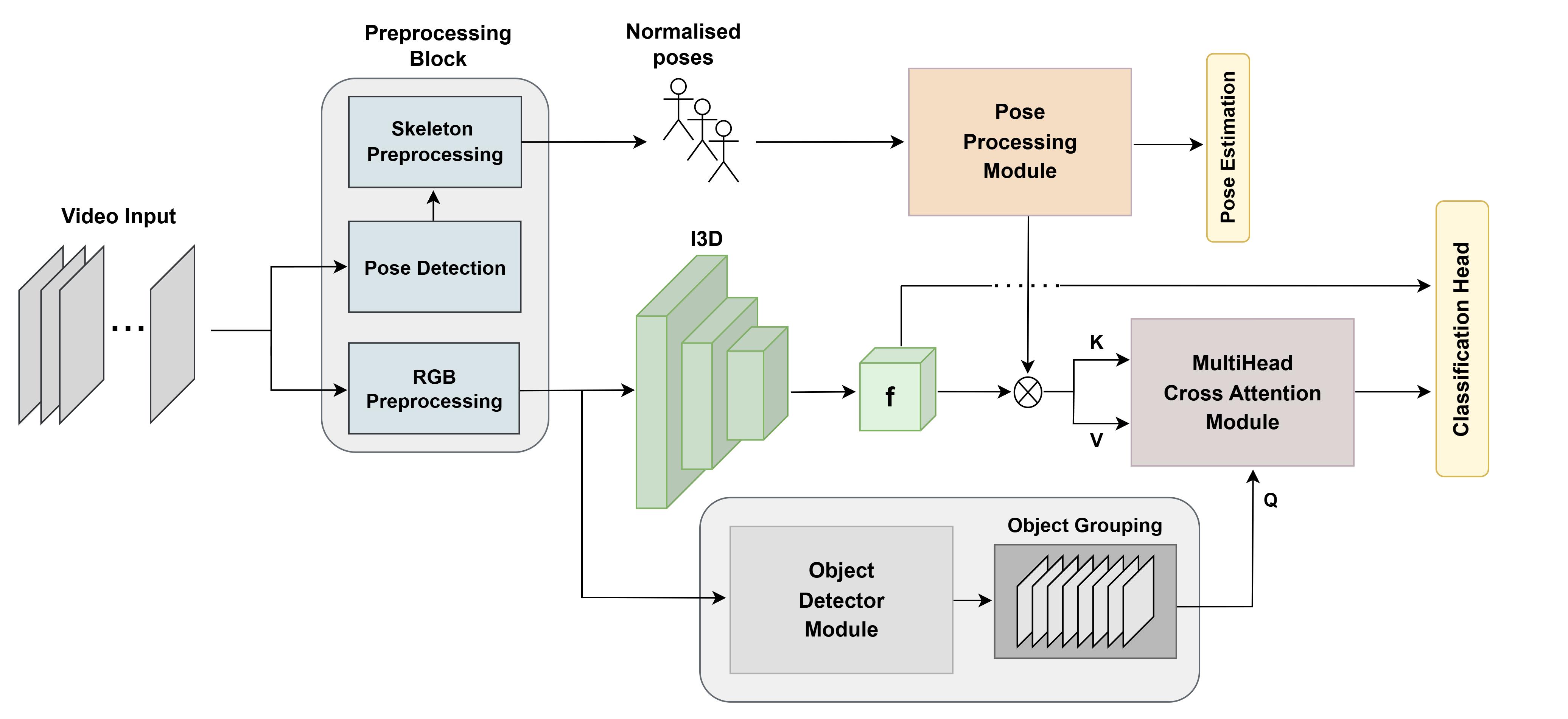}
    \caption{Architecture of the Proposed Method}
    \label{fig:architecture}
\end{figure}

The overall architecture is illustrated in Figure~\ref{fig:architecture} and consists of four main components: (1) data preprocessing for video and pose streams, (2) feature extraction networks for each modality, (3) a cross-attention fusion module, and (4) a classification head for final activity prediction. Details for each component are provided below.

\subsection{Video and Pose Data Preprocessing}

To address the challenges of view variance and environmental variability in indoor activity recognition, preprocessing techniques are applied to normalise both skeletal and video data streams. These steps, inspired by~\citep{climent2021improved}, aim to reduce intra-class variability while preserving the essential characteristics required for robust activity classification.

\begin{figure}[t]
    \centering
    \begin{subfigure}[t]{0.4\textwidth}
        \centering
        \includegraphics[width=\linewidth]{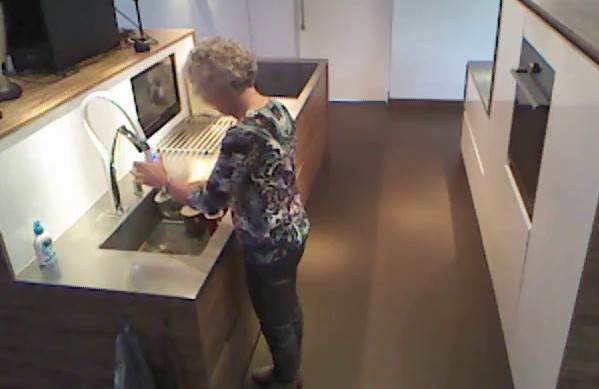}
        \caption{Original RGB frame}
    \end{subfigure}
    \hfill
    \begin{subfigure}[t]{0.55\textwidth}
        \centering
        \includegraphics[width=\linewidth]{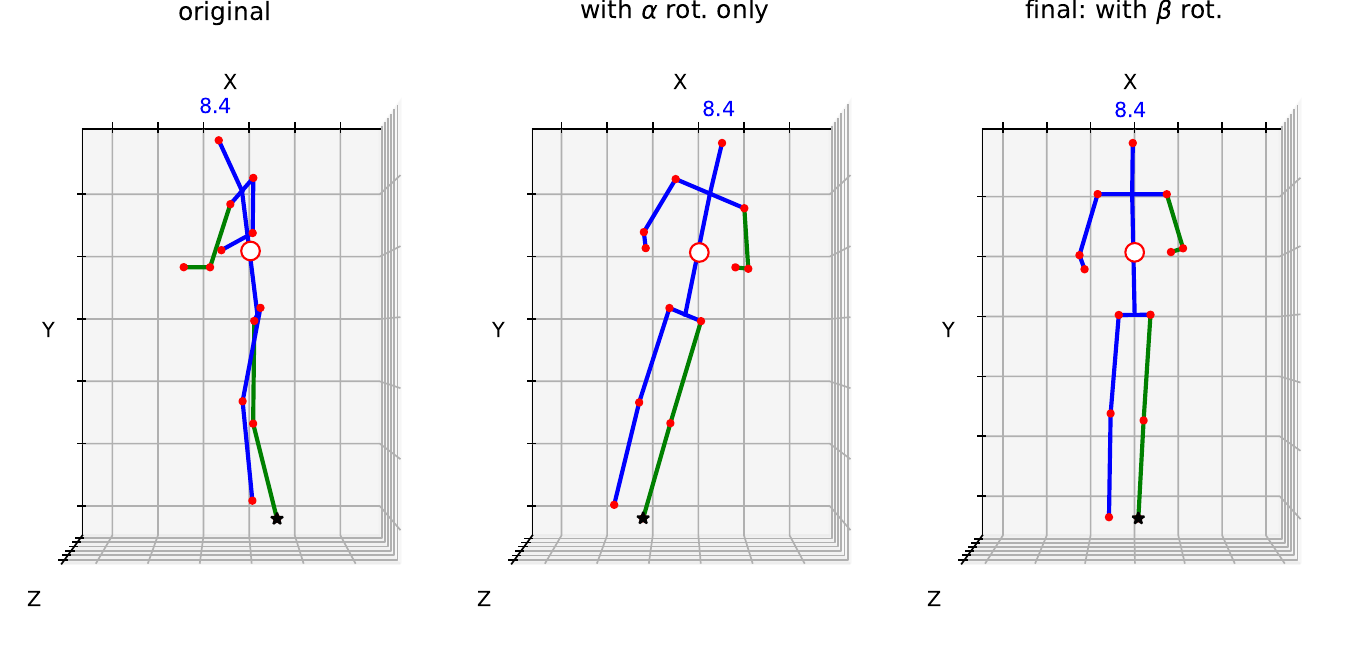}
        \caption{Skeleton plots}
    \end{subfigure}
   \caption{Sequential skeleton rotation stages adapted from~\citep{climent2021improved}. (\textbf{a}) Original frame; (\textbf{b}) left: base pose, center: $Y$-axis rotated pose to face forward, right: fully rotated pose ($Y$ and $Z$ axes).}
    \label{fig:skeleton_rotation} 
\end{figure}

The 3D human pose data undergoes a two-stage rotation process to create view-invariant skeletal representations (Figure~\ref{fig:skeleton_rotation}). This normalization is crucial for indoor HAR systems where the same activity may be captured from different camera perspectives, leading to unnecessary variations in the skeletal joint positions. First, a Y-axis rotation is applied to ensure all skeletons are oriented to "face forward" regardless of the original camera viewpoint. The rotation angle $\alpha$ is calculated using three key joints - left shoulder ($\vec{s_l}$), right shoulder ($\vec{s_r}$), and right hip ($\vec{h_r}$) - which define the torso plane:

\begin{equation}
\alpha = \arctan\left(\frac{s_{lz} - \frac{s_{rz} + h_{rz}}{2}}{s_{lx} - \frac{s_{rx} + h_{rx}}{2}}\right)
\end{equation}

This rotation is computed once at the beginning of each sequence ($t_0$) to maintain natural body rotations that occur as part of the activity itself. Second, a Z-axis rotation compensates for camera tilt with respect to the ground plane. This rotation $\beta$ is calculated at each frame to account for temporal variations in pose estimation:

\begin{equation}
\beta_t = \frac{\beta_{s,t} + \beta_{h,t}}{2}
\end{equation}

\noindent where $\beta_{s,t}$ and $\beta_{h,t}$ represent independent angle estimations from shoulder and hip joint pairs respectively. This frame-by-frame correction ensures consistent skeletal orientation throughout the activity sequence.

Complementing the skeletal normalization, the RGB video preprocessing employs a "full activity crop" strategy that captures the complete spatial context of human activities. Compared to only cropping around an individual person, this method considers the entire spatial footprint of the activity over time. The full activity crop is defined by identifying the minimum and maximum coordinates of all human detections throughout the video sequence:

\begin{equation}
\vec{p_{TL}} = \min_{i=1..t}(x_i, y_i), \quad \vec{p_{BR}} = \max_{i=1..t}(x_i, y_i)
\end{equation}

This creates a bounding box from the top-left ($\vec{p_{TL}}$) to bottom-right ($\vec{p_{BR}}$) corners, that encompasses the complete activity space, which is particularly beneficial for activities where spatial displacement is a key discriminative feature (e.g., walking, moving between locations). Since the neural network architecture requires a square input, a square crop is generated by finding the center of the activity bounding box and using the larger dimension as the square side length. When the resulting crop extends beyond the image boundaries, padding with neutral gray values (RGB: 128) maintains the aspect ratio while preserving the spatial relationships within the activity region.

This preprocessing approach addresses two key limitations: it provides more accurate person detection through improved bounding box estimation, and it preserves the spatial dynamics of activities that may be lost when cropping around individual frame detections. The combination of view-invariant skeletal normalization and comprehensive spatial context preservation in video data creates a more robust foundation for multi-modal activity recognition in indoor environments.

\subsection{Feature Learning Components}
Following the preprocessing stage, each modality undergoes specialized feature extraction to capture its unique characteristics. The visual stream employs a 3D Convolutional Neural Network (I3D)~\citep{carreira2017quo} to extract spatio-temporal features from the normalized video data, while the pose stream utilizes GCN to model pose information and temporal dynamics in the skeletal data. The object detection component provides semantic context by identifying relevant objects within the activity region. Additionally, a pre-trained object detection module identifies and localizes relevant objects within the activity region to provide contextual information for attention-based fusion.

\subsubsection{Visual Feature Extraction with I3D}
The RGB video component of our multi-modal framework employs I3D as the feature extraction backbone for processing the preprocessed RGB video data. Following the preprocessing stage, the input consists of temporally-ordered human activity crops with dimensions corresponding to the full activity spatial extent, determined during the preprocessing phase.

The I3D network processes sequences of $T$ sequential frames, sampled from the preprocessed video $V$, where each frame contains the square-cropped activity region. The spatio-temporal representation $\mathbf{f}$ is extracted from an intermediate layer of the I3D network, specifically the convolutional layer preceding the Global Average Pooling (GAP) operation. This choice preserves both spatial and temporal granularity necessary for subsequent attention-based fusion with pose and object context information.

The resulting feature map $\mathbf{f}$ has $t \times h \times w \times c$ dimensions, where $t$ represents the temporal dimension, $h \times w$ corresponds to the spatial resolution of the feature map, and $c$ denotes the number of feature channels. This 4-dimensional representation captures hierarchical spatio-temporal patterns that encode both local motion dynamics and global activity structure, making it suitable for distinguishing between activities with similar motion patterns but different spatial contexts~\citep{carreira2017quo}.

The I3D backbone uses pre-trained weights from large-scale action recognition datasets, providing robust feature representations that transfer effectively to the indoor AAL domain. The extracted features $\mathbf{f}$ serve as input to the cross-attention mechanism, where they are spatially and temporally modulated based on complementary information from the pose and object detection modalities.

\subsubsection{Pose Feature Extraction with GCN}
To address the view variance challenges inherent in indoor AAL environments, our framework incorporates a specialized pose analysis component that leverages the natural structural properties of human skeletal data. The skeletal modality provides complementary geometric information that remains consistent across different camera viewpoints, making it particularly valuable for robust activity recognition in home monitoring scenarios.

Human pose sequences are represented as a temporal collection of 3D joint coordinates $P \in \mathbb{R}^{3 \times J \times T}$, where each frame $P_t \in \mathbb{R}^{3 \times J}$ contains the spatial positions of $J$ anatomical joints. Rather than treating these coordinates as independent features, we exploit the inherent connectivity structure of the human body by modeling each pose frame as an undirected graph where joints serve as vertices and anatomical connections define the edge relationships.

The graph representation $G_t(P_t, A)$ at each temporal step $t$ incorporates a weighted adjacency matrix $A \in \mathbb{R}^{J \times J}$ that encodes the natural biomechanical relationships between body parts:

\begin{equation}
A_{ij} = \begin{cases}
0, & \text{if } i = j \\
\alpha, & \text{if joint } i \text{ and joint } j \text{ are connected} \\
\beta, & \text{if joint } i \text{ and joint } j \text{ are disconnected}

\end{cases}
\end{equation}

This weighted formulation enables the network to learn differentiated importance between directly connected joints (e.g., shoulder-elbow) and distant but semantically related joints (e.g., hand-foot coordination patterns).

The graph convolution operation processes these structured pose representations through learnable transformations that respect the skeletal topology. For each temporal frame, the GCN applies normalized spectral convolution to propagate information between connected joints:

\begin{equation}
H_t^{(l+1)} = D^{-\frac{1}{2}}(A + I)D^{-\frac{1}{2}}P_t^T W^{(l)},
\end{equation}

\noindent where the degree normalization matrix $D$ ensures stable information flow across the graph structure, $P_t^T \in \mathbb{R}^{J \times 3}$ represents the pose coordinates in standard node-feature format, and $W^{(l)}$ contains learnable parameters for layer $l$. The inclusion of self-connections through the identity matrix $I$ allows each joint to retain its original positional information while incorporating contextual information from neighboring joints.

The temporal dimension is handled by independently processing each frame $t \in \{1, 2, ..., T\}$ through the graph convolution layers, resulting in a sequence of enhanced joint representations $\{H_1^{(l+1)}, H_2^{(l+1)}, ..., H_T^{(l+1)}\}$. These temporally-distributed features capture both the instantaneous geometric configuration of the human pose and the spatial relationships that define body part coordination.

\begin{figure}[t]
    \centering
    \includegraphics[width=\textwidth]{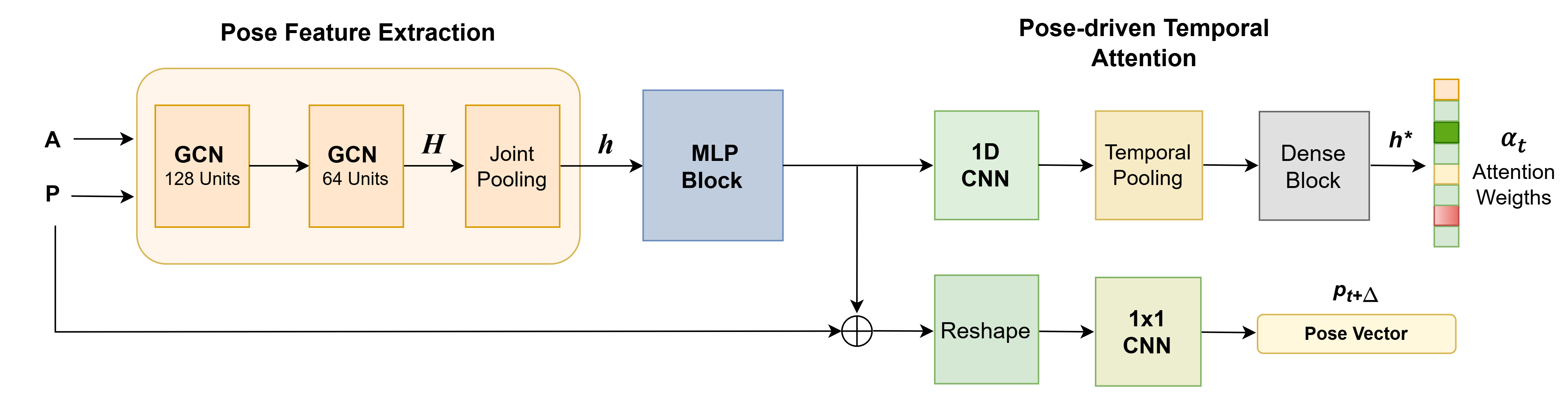}
    \caption{Pose processing module with two tasks, calculating temporal attention weights and pose estimation.}
    \label{fig:pose}
\end{figure}

As illustrated in Figure~\ref{fig:pose}, following graph convolution processing, a residual connection preserves the original pose information while allowing the network to learn additive improvements through structural modeling. Additional convolution operations further refine these enhanced representations to produce the final pose features $h^*$, which encode view-invariant geometric patterns suitable for guiding temporal attention mechanisms.

\subsubsection{Object Detection Module}
To capture the contextual information essential for distinguishing between activities with similar motion patterns, we incorporate a pre-trained object detection module that identifies and localizes relevant objects within the activity region. The object detection component provides semantic context by recognizing household items, furniture, and manipulated objects that define the activity context, addressing the object interaction complexity challenge inherent in AAL environments.

For each video sequence, all frames are processed to identify and localize relevant objects. Rather than using frame-level detections independently, temporal object masks are created that account for object movement throughout the video sequence. For each detected object type, a unified spatial mask $M_o^{video}$ is generated by computing the union of all bounding box regions where the object appears across all cropped frames:

\begin{equation}
M_o^{video} = \bigcup_{t=1}^{T} M_o^{(t)}
\end{equation}

\noindent where $M_o^{(t)}$ represents the binary mask of object $o$ at frame $t$, and $T$ is the total number of frames in the video sequence. This temporal aggregation strategy ensures that the spatial attention mechanism can focus on regions where objects are active throughout the entire activity sequence, rather than being limited to instantaneous object positions.

To maintain computational efficiency while preserving semantic richness, the dimensionality of object representations is reduced through strategic grouping based on their co-occurrence patterns in indoor scenes. Objects that rarely appear together in the same activity context are grouped together, minimizing information loss while reducing computational overhead for the cross-attention mechanism (see Section~\ref{lab:object-grouping}).

For each group, a consolidated spatial mask $M_g^{video}$ is generated by computing the union of all individual object masks within that group:

\begin{equation}
M_g^{video} = \bigcup_{o \in G_g} M_o^{video}
\end{equation}

\noindent where $G_g$ represents the set of objects belonging to group $g$. This produces group-level spatial masks per video that capture the spatial distribution of semantically related objects while maintaining computational tractability. Each group-level spatial mask is resized to match the spatial dimensions $w \times h$ of the I3D feature output, enabling their direct use in creating queries in the cross-attention mechanism. This allows the model to selectively focus on spatial regions where relevant objects are present during activity execution. The object context information guides the spatial attention weights applied to the visual feature maps $\mathbf{f}$ extracted from the I3D network, highlighting the object interaction regions crucial to distinguishing between activities with similar motion patterns but different contextual elements.

The integration of object detection with spatial attention mechanisms addresses the fundamental challenge that many indoor activities are defined by the manipulation of specific objects rather than just human motion patterns. By providing explicit object context through spatial masks, the discriminative power of the multi-modal fusion process is enhanced, enabling more accurate fine-grained activity recognition in complex indoor environments.

\subsection{Modality Fusion and Classification}

To effectively integrate complementary information from human pose and object context modalities with the spatio-temporal visual features extracted by the 3D CNN, a two-stage attention-based fusion mechanism is employed. This design enables dynamic weighting of temporal video features based on pose-derived activity relevance, followed by spatial refinement guided by object interaction context. The resulting fused representation enhances discriminative capacity for fine-grained daily activity recognition in indoor AAL environments.

\subsubsection{Pose-Driven Temporal Attention}

Human pose sequences inherently capture view-invariant geometric and dynamic patterns relevant to activity progression. To leverage this modality for temporally adaptive feature weighting, the output of the pose processing GCN is first aggregated and transformed to produce a temporal attention vector that modulates the visual feature map along the temporal dimension.

Specifically, as shown in Figure~\ref{fig:pose}, the pose feature tensor $H \in \mathbb{R}^{T_p \times J \times C_p}$, where $T_p$ is the input temporal length, $J$ is the number of joints, and $C_p$ is the dimension of the pose feature channel, is spatially pooled over the dimension of the joint to obtain temporally ordered pose embeddings:

\begin{equation}
h_t = \frac{1}{J} \sum_{j=1}^{J} H_{t,j,:} \in \mathbb{R}^{C_p}, \quad t = 1, \ldots, T_p
\end{equation}

The sequence $\{h_t\}_{t=1}^{T_p}$ is passed through a multi-layer perceptron (MLP) block
producing an output tensor of shape $T_p \times (3 \times J)$ to match the dimensions of the original pose tensor $P$. This dimensional alignment enables a residual connection between the predicted and original pose tensors. 
This output is then reshaped to $(J \times 3 \times T_p)$ and processed by a 1×1 convolutional layer that reduces the temporal dimension to 1: $(J \times 3 \times T_p) \rightarrow (J \times 3 \times 1)$, to predict the future pose at time $t + \Delta$, where each prediction yields a pose vector in $\mathbb{R}^{3 \times J}$, constituting \textbf{an auxiliary pose estimation task}. This auxiliary task encourages the pose features to capture meaningful temporal dynamics by learning to forecast subsequent human joint configurations, thereby enhancing the semantic relevance of the temporal attention weights applied to the visual features.

To compute temporal attention weights, from the shared MLP layer, a lightweight 1D convolutional layer is first applied to enhance the temporal features. Then,
the temporally refined pose features $h^*$ are obtained by first applying temporal pooling
that averages every $n~=~T_p / T$ frames, where $T$ is the temporal dimension of the I3D    visual feature map $\mathbf{f} \in \mathbb{R}^{T \times H \times W \times C}$, followed by a subsequent Dense block. 
The pooling operation reduces the sequence length from $T_p$ to $T$ to match the temporal dimension of the visual features. 
These refined features are then used to derive a temporal attention vector $\alpha \in \mathbb{R}^T$ by normalizing $h^*$ via a softmax function to produce a distribution of importance coefficients over the temporal dimension:

\begin{equation}
\alpha_t = \frac{\exp(h^*_t)}{\sum_{k=1}^T \exp(h^*_k)}, \quad t=1,\ldots,T
\end{equation}

This attention vector $\boldsymbol{\alpha}$ is then applied multiplicatively across the temporal slices of the video feature map $\mathbf{f}$, modulating the contribution of each temporal frame according to pose-derived activity relevance:

\begin{equation}
\tilde{\mathbf{f}}_{t,:,:,:} = \alpha_t \cdot \mathbf{f}_{t,:,:,:}, \quad t=1,\ldots,T
\end{equation}

By applying pose-driven temporal attention prior to spatial fusion, the framework dynamically emphasizes temporal segments of the video that are more informative for the activity, guided by the intrinsic understanding of the pose modality of human motion dynamics.

\subsubsection{Object-Guided Spatial Cross-Attention}

Following temporal weighting, spatial contextualization is performed using grouped object masks derived from the object detection module. These masks serve as semantic queries attending over the temporally modulated visual feature map $\tilde{\mathbf{f}}$, enabling selective focus on spatial regions corresponding to relevant object interactions.

The 4D feature map $\tilde{\mathbf{f}} \in \mathbb{R}^{T \times H \times W \times C}$ is reshaped into a flattened spatio-temporal token sequence:

\begin{equation}
\mathbf{F} \in \mathbb{R}^{(T \cdot H \cdot W) \times C}
\end{equation}

For each set of the $G$ object groups, a spatial mask is generated by aggregating detections over time and resizing it to match the spatial dimensions of the feature map, resulting in $M_g \in \mathbb{R}^{H \times W}$. The mask is then replicated across the $T$ temporal frames to align with $\tilde{\mathbf{f}}$. Masked average pooling over the spatio-temporal features produces group-level query vectors:

\begin{equation}
\mathbf{q}_g = \frac{1}{\sum_{i,j} M_g(i,j) \cdot T} \sum_{t=1}^T \sum_{i=1}^H \sum_{j=1}^W M_g(i,j) \cdot \tilde{\mathbf{f}}_{t,i,j,:} \in \mathbb{R}^C
\end{equation}

Each output remains a vector in $\mathbb{R}^C$, representing a channel-wise summary of the salient features of the corresponding object group. In other words, encoding "what is important" for that group, aggregated over the entire video sequence. Stacking all group queries yields the query matrix:

\begin{equation}
\mathbf{Q} = [\mathbf{q}_1, \mathbf{q}_2, \ldots, \mathbf{q}_G]^T \in \mathbb{R}^{G \times C}
\end{equation}

The key and value matrices are set as follows:

\begin{equation}
\mathbf{K} = \mathbf{V} = \mathbf{F} \in \mathbb{R}^{(T \cdot H \cdot W) \times C}
\end{equation}

\noindent and multi-head cross-attention is then applied:

\begin{equation}
\mathbf{A} = \text{MultiHeadCrossAttention}(\mathbf{Q}, \mathbf{K}, \mathbf{V}) \in \mathbb{R}^{G \times C}
\end{equation}

\noindent where each output vector $\mathbf{a}_g$ encapsulates a contextually attended summary of visual features relevant to object group $g$.

\begin{figure}[t]
    \centering
    \includegraphics[width=0.95\textwidth]{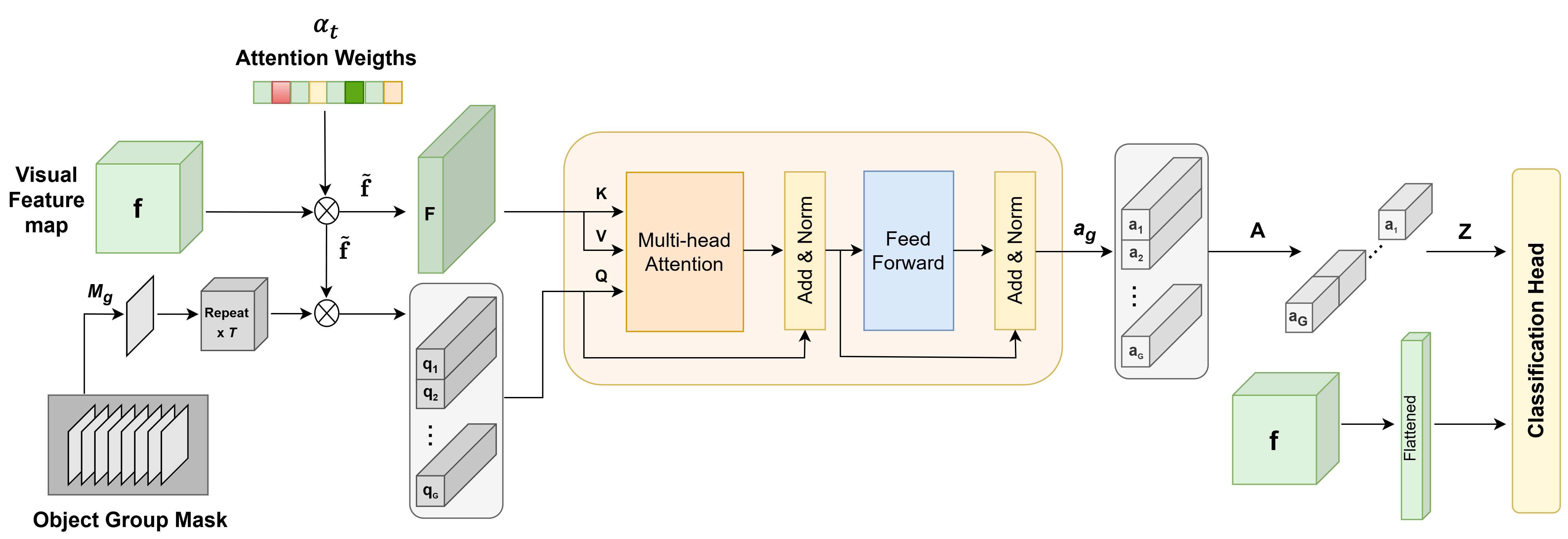}
    \caption{Cross-Attention mechanism between scene context and extracted visual features.}
    \label{fig:att}
\end{figure}

\subsubsection{Fusion and Final Classification}

The attended outputs from all object groups are concatenated to form a unified contextual embedding:

\begin{equation}
\mathbf{z}_{context} = [\mathbf{a}_1 \| \mathbf{a}_2 \| \ldots \| \mathbf{a}_G] \in \mathbb{R}^{G \cdot C}
\end{equation}

This embedding is combined with the temporally pooled visual feature vector, obtained by global average pooling over $\tilde{\mathbf{f}}$, and concatenated with pose features pooled over time, to form a comprehensive multi-modal representation.

The fused vector is then passed through fully connected layers with nonlinear activations and dropout regularization before reaching the final softmax classification layer, which outputs the activity class probabilities. Figure~\ref{fig:att} summarizes the process from the attention mechanism to the classification head.

\subsubsection{Multi-Task Loss Function}

The network is trained using a weighted multi-task loss function that combines the primary activity classification objective with the auxiliary pose estimation task. The total loss is formulated as:

\begin{equation}
L_{total} = L_{act} + \lambda_{pose} \cdot L_{pose}
\end{equation}

\noindent where $L_{act}$ represents the cross-entropy loss for the main activity classification task, and $L_{pose}$ denotes the loss for the auxiliary pose prediction task. The weighting parameter $\lambda_{pose}$ is tunable and controls the influence of the pose classification task on the overall training objective.

The main classification loss is computed as:

\begin{equation}
L_{act} = -\frac{1}{N} \sum_{i=1}^{N} \sum_{c=1}^{C} y_{i,c} \log(\hat{y}_{i,c})
\end{equation}

\noindent where $N$ is the batch size, $C$ is the number of activity classes, $y_{i,c}$ is the ground truth label, and $\hat{y}_{i,c}$ is the predicted probability for sample $i$ and class $c$.

The auxiliary pose loss encourages the temporal attention mechanism to learn meaningful pose dynamics by predicting future joint configurations:

\begin{equation}
L_{pose} = \frac{1}{N} \sum_{i=1}^{N} \|\hat{p}_{i,t+\Delta} - p_{i,t+\Delta}\|_2^2
\end{equation}

\noindent where $\hat{p}_{i,t+\Delta}$ is the predicted pose at time $t+\Delta$ and $p_{i,t+\Delta}$ is the ground truth pose. This multi-task learning approach ensures that the pose-driven temporal attention weights are semantically meaningful and aligned with human motion dynamics, thereby improving the overall activity recognition performance.

\subsubsection{Discussion on Design Choices}

The sequential application of pose-driven temporal attention followed by object-guided spatial cross-attention enables the model to first emphasize temporally relevant frames based on intrinsic human motion cues, then spatially refine these frames using contextual object information. This order is justified because temporal weighting benefits from the direct insight of the pose modality into motion dynamics, while spatial attention uses the explicit object layout to discriminate visually similar actions involving different object interactions.

The auxiliary pose classification head ensures that the learned temporal attention is semantically meaningful and aligned with the activity recognition objective, mitigating the risk of spurious or noisy attention weights. This multi-task learning approach contributes to improved convergence and robustness.

Grouping objects into semantically distinct clusters reduces computational overhead and enforces structured attention, allowing the model to generalize across diverse indoor scenes without overfitting to individual object instances.

The architectural decision to provide residual connections only to the auxiliary prediction branch, while excluding direct pose access from the temporal attention mechanism, serves two important purposes. First, this separation enables specialized final processing where each branch can adapt its final representation to its specific objective, while still benefiting from shared temporal feature learning that improves both tasks; the auxiliary prediction benefits from direct pose residuals for future pose estimation, while the attention branch can focus on learning higher-level temporal patterns. Second, this design prevents information leakage where the temporal attention might become overly dependent on direct pose coordinates rather than learning meaningful temporal patterns that generalize across different pose configurations and viewpoints. This approach ensures that the auxiliary pose prediction task fulfills its primary role of enhancing the shared representations for the main activity recognition objective, while allowing each branch to specialize in its final processing steps.

Overall, the proposed fusion mechanism addresses key challenges in indoor activity recognition (temporal variability, object interaction complexity, and inter-class similarity) by leveraging complementary modalities through principled attention operations.

\section{Experimental Setup}
\label{sec:experimental_setup}

\subsection{Dataset}
Experiments have been conducted on the Toyota SmartHome dataset~\citep{das2019toyota}, which comprises 16,115 video clips of daily living activities performed by 18 senior participants aged 60-80 years. The dataset provides multi-modal data including RGB video (original resolution 640×480), depth information, and 3D skeletal pose data captured using seven Kinect v1 cameras positioned across three indoor scenes: dining room, living room, and kitchen. All activities were fully unscripted with participants unaware of the study purpose, ensuring naturalistic behavior patterns. To the best of our knowledge, this is the only dataset that is aligned well with our focus on older adults' daily activities in realistic home environments for AAL applications.

\subsection{Pre-processing and Temporal Alignment}
To handle the multi-modal temporal alignment, input dimensions of 224×224 pixels are utilized after human-centered cropping, with the I3D network processing 64-frame sequences while the GCN processes 32-frame pose sequences using 13 joints with 3D coordinates. Temporal alignment between modalities is achieved through uniform sampling, where pose frames are sampled at every second frame to match the temporal coverage of the video sequence. 

\subsection{Backbones and Features}
The I3D network with Inception backbone utilizes pre-trained weights from the Kinetics-400 dataset, with features extracted from the 8×7×7×1024 dimensional layer before global average pooling. The GCN employs two graph convolution layers processing each pose frame independently, with hidden dimensions of 128 and 64 units respectively, adjacency matrix weights $\alpha=5$ for connected joints and $\beta=2$ for disconnected joints, and ReLU activation functions. 

\subsection{Object Detection and Grouping}
\label{lab:object-grouping}
Object detection is performed using YOLOv8 medium variant trained on the COCO dataset with confidence threshold of 0.5. From the COCO vocabulary we retain 40 home-relevant classes. A naïve design would allocate one spatial mask per object (40 masks), but this inflates memory/compute in the fusion module and increases supervision ambiguity when multiple masks activate simultaneously. To control complexity, we reduce the masks to eight by grouping objects. We first considered room-based grouping (e.g., ``kitchen''), but this placed many co-activating items into the same mask (e.g., sink, microwave, knife), which hindered discrimination between distinct actions occurring within the same room. Instead, we adopt a few-coincidences strategy that merges objects that rarely appear in the same activities, yielding eight masks that are less likely to co-activate while preserving informative spatial cues. This grouping process is as follows:

Let $\mathcal{V}$ be the set of videos and $\mathcal{A}$ the set of activities. For each object $o_i$ and video $v$, define the object--video incidence
\[
Z_{i,v} \in \{0,1\}, \quad Z_{i,v}=1 \iff o_i \text{ is detected at least once in } v.
\]
For each activity $a\in\mathcal{A}$, let $\mathcal{V}_a\subset \mathcal{V}$ be the videos of activity $a$. We form activity-wise object counts
\[
S_{a,i} \;=\; \sum_{v\in \mathcal{V}_a} Z_{i,v},
\]
so row $a$ of $S$ describes how often object $i$ appears in activity $a$.

To compare objects independent of their marginal frequency, we normalise each object column over activities:
\[
P_{a,i} \;=\; \frac{S_{a,i}}{\sum_{a' \in \mathcal{A}} S_{a',i}} \quad.
\]
Column $P_{\cdot,i}$ is the empirical distribution of object $i$ across activities.

We measure how differently two objects distribute over activities using the Pearson correlation of their columns, $\mathrm{corr}(P_{\cdot,i},P_{\cdot,j})$. Low (or negative) correlation indicates few coincidences across activities. 

We start with 40 singleton groups, one for each object. At each step we merge the pair $(G_p,G_q)$ with the smallest correlation. For any group $G$, we define group–video presence by an $n$-ary OR
\[
Z_{G,v} \;=\; \bigvee_{o_i \in G} Z_{i,v},
\]
and rebuild $S$ and $P$ at the group level: $S_{a,G}=\sum_{v\in\mathcal{V}_a} Z_{G,v}$ and
\[
P_{a,G}\;=\;\frac{S_{a,G}}{\sum_{a' \in \mathcal{A}} S_{a',G}}.
\]
We then recompute the correlation matrix over the current groups and repeat until eight groups remain. For reproducibility, the resulting groups are:
\begin{itemize}
  \item \textbf{G1:} Knife, Vase, Hair Drier
  \item \textbf{G2:} Sink, Orange, Dining Table
  \item \textbf{G3:} Remote, Pizza, Mouse, Broccoli, Toaster
  \item \textbf{G4:} Refrigerator, Cup, Cake, Microwave
  \item \textbf{G5:} TV, Hot Dog, Fork, Toilet, Wine Glass, Donuts, Bed, Toothbrush
  \item \textbf{G6:} Bench, Banana, Couch, Spoon, Apple
  \item \textbf{G7:} Chair, Bowl, Laptop
  \item \textbf{G8:} Book, Sandwich, Keyboard, Bottle, Carrot, Cell Phone, Oven, Scissors
\end{itemize}

\subsection{Fusion}
For each clip we compute one binary mask per group by the temporal union of its member-object masks. The cross-attention fusion uses these 8 masks as queries and employs 8 attention heads to capture diverse object-guided spatial patterns.

\subsection{Training Details}
The network is trained using the Adam optimizer with a learning rate of 0.001, batch size of 16, and 100 epochs. Dropout regularization with rate 0.3 is applied to prevent overfitting, and early stopping is used with patience of 10 epochs based on validation accuracy. The multi-task loss weighting parameter $\lambda_{pose}$ is set to 0.25, with pose prediction horizon of 3 frames ahead for the auxiliary task. Training is carried out on a Tesla A100 GPU with 40GB memory using TensorFlow framework, requiring approximately 30 hours of training time.

\subsection{Evaluation Protocols}
We follow the original Cross-Subject (CS) and Cross-View (CV) protocols~\citep{das2019toyota} for evaluation. In the CS setting, 11 subjects are used for training and 7 for testing across 35 activity classes, including drinking, cooking, movement, and daily living tasks. In the CV setting, the CV1 and CV2 protocols are adopted with 19 activity classes, where different camera combinations are used for training, validation, and testing to assess view-invariant performance.

Performance is measured using mean per-class accuracy, following the established Toyota SmartHome evaluation procedures. Results are reported separately for the CS, CV1, and CV2 scenarios. To validate the effectiveness of the proposed multi-modal fusion approach, comparisons are made against various single- and multi-modal baselines.

\section{ Results and Discussion}\label{sec5}

This section presents a comprehensive evaluation of the proposed multi–modal daily activity recognition framework on the Toyota SmartHome dataset, following the CS, CV1, and CV2 evaluation protocols described in Section~\ref{sec:experimental_setup}. We first report the overall performance of the proposed method in comparison with representative single- and multi-modal approaches. This is followed by detailed ablation studies analyzing the contribution of each modality and fusion component, qualitative visualization of the learned attention mechanisms, and an error analysis highlighting the strengths and limitations of the method.

\subsection{Overall Performance}

\begin{table}[t]
\centering
\caption{Comparison of mean per-class accuracy (\%) on Toyota SmartHome dataset across different evaluation protocols. $\circ$ indicates that the modality has been used only in training.}
\label{tab:comparison_results}
\begin{tabular}{p{5cm}@{\hspace{1cm}}c@{\hspace{1cm}}c@{\hspace{1cm}}c@{\hspace{1cm}}c@{\hspace{1cm}}c}
\textbf{Methods} & \textbf{Pose} & \textbf{RGB} & \textbf{CS} & \textbf{CV1} & \textbf{CV2} \\
\hline
\multicolumn{6}{l}{\textbf{\textit{Pose Only}}} \\
2s-AGCN~\citep{shi2019two} & $\checkmark$ & $\times$ & 60.9 & 21.6 & 32.3 \\
ST-GCN~\citep{yan2018spatial} & $\checkmark$ & $\times$ & 53.8 & 12.5 & 51.0 \\
UNIK~\citep{yang2021unik} & $\checkmark$ & $\times$ & 63.1 & 22.9 & 61.4 \\
MS-G3D Net~\citep{liu2020disentangling} & $\checkmark$ & $\times$ & 66.7 & 17.5 & 59.4 \\
ViA~\citep{yang2024view} & $\checkmark$ & $\times$ & 64.0 & 35.6 & 65.4 \\
\hline
\multicolumn{6}{l}{\textbf{\textit{RGB Only}}} \\
I3D~\citep{carreira2017quo} & $\times$ & $\checkmark$ & 53.4 & 34.9 & 45.1 \\
AssembleNet++~\citep{ryoo2020assemblenet++} & $\times$ & $\checkmark$ & 63.6 & -- & -- \\
LTN~\citep{yang2023self} & $\times$ & $\checkmark$ & 65.9 & -- & 54.6 \\
VPN++~\citep{das2021vpn++} & $\circ$ & $\checkmark$ & 69.0 & -- & 54.9 \\
$\pi$-ViT~\citep{reilly2024just} & $\circ$ & $\checkmark$ & 72.9 & 55.2 & 64.8 \\
SV-data2vec~\citep{dovzdor2025sv} & $\circ$ & $\checkmark$ & 72.9 & 46.0 & 57.5 \\
\hline
\multicolumn{6}{l}{\textbf{\textit{RGB + Pose}}} \\
P-I3D~\citep{das2019focus} & $\checkmark$ & $\checkmark$ & 54.2 & 35.1 & 50.3 \\
Separable STA~\citep{das2019toyota} & $\checkmark$ & $\checkmark$ & 54.2 & 35.2 & 50.3 \\
VPN~\citep{das2020vpn} & $\checkmark$ & $\checkmark$ & 65.2 & 43.8 & 54.1 \\
VPN++ + 3D Poses~\citep{das2021vpn++} & $\checkmark$ & $\checkmark$ & 71.0 & -- & 58.1 \\
MMNet~\citep{bruce2022mmnet} & $\checkmark$ & $\checkmark$ & 70.1 & 37.4 & 46.6 \\
$\pi$-ViT + 3D Poses~\citep{reilly2024just} & $\checkmark$ & $\checkmark$ & 73.1 & 55.6 & 65.0 \\
\textbf{Ours} & $\checkmark$ & $\checkmark$ & \textbf{70.1} & \textbf{44.2} & \textbf{65.4} \\
\hline
\end{tabular}
\end{table}

Table~\ref{tab:comparison_results} summarizes the mean per-class accuracy of the proposed system and the compared baseline methods under the three evaluation protocols. The results indicate that the proposed framework consistently outperforms the individual single-modal baselines (video-only, pose-only) as well as conventional fusion strategies, demonstrating the 
advantage of the proposed method.

Under the CS protocol, our approach achieves 70.1\% accuracy, which is competitive with recent transformer-based state-of-the-art methods while using much lighter architecture, providing practical implementation advantages. While $\pi$-ViT achieves superior performance (72.9\% CS), it employs a heavy transformer-based architecture with TimeSformer backbone, requires large-scale pretraining dataset, and utilizes dual auxiliary modules (2D-SIM and 3D-SIM) with separate token-skeleton mapping and feature alignment tasks. Similarly, SV-data2vec (72.9\% CS) employs a computationally intensive transformer architecture with self-supervised pretraining using teacher-student framework requiring exponential moving average updates and complex masked prediction training procedures.

Remarkably, our multi-modal architecture demonstrates that domain-tailored computationally efficient methods can effectively bridge the performance gap with resource-intensive transformer architectures while maintaining significant computational advantages. Furthermore, within the CNN-GCN architectural category, our approach significantly outperforms existing methods, including P-I3D, Separable STA, and VPN, by substantial margins in all the protocols, confirming superior fusion design. Notably, on the CV2 protocol, our method exhibits superior view-invariant capabilities (65.4\%), surpassing both $\pi$-ViT (64.8\%) and SV-data2vec (57.5\%), thus rivaling transformer-based approaches in challenging cross-view scenarios.

\subsection{System Component Evaluation}
To validate the effectiveness of each component in the proposed multi-modal framework, a comprehensive study on the Toyota SmartHome dataset is conducted across all three evaluation protocols. This analysis examines the contribution of individual modalities, fusion strategies, preprocessing techniques, and architectural design choices.

\begin{table}[t]
\centering
\caption{Comprehensive ablation study results showing mean per-class accuracy (\%) on Toyota SmartHome dataset across different evaluation protocols.}
\label{tab:ablation_results}
\begin{tabular}{p{8cm}@{\hspace{.8cm}}c@{\hspace{.8cm}}c@{\hspace{.8cm}}c}
\textbf{Configuration} & \textbf{CS} & \textbf{CV1} & \textbf{CV2} \\
\hline
\multicolumn{4}{l}{\textbf{\textit{Single-Modal Baselines}}} \\
Video-only (I3D) & 53.4 & 34.9 & 45.1 \\
Pose-only (GCNs) & 52.1 & 42.6 & 51.2 \\
\hline
\multicolumn{4}{l}{\textbf{\textit{Bi-Modal Baselines}}} \\
Video + Pose & 68.4 & 42.7 & 62.1 \\
Video + Object & 66.8 & 39.5 & 60.3 \\
\hline
\multicolumn{4}{l}{\textbf{\textit{Multi-Modal Baselines}}} \\
\textbf{Full system} & \textbf{70.1} & \textbf{44.2} & \textbf{65.4} \\
w/o pose normalization & 67.8 & 40.1 & 62.9 \\
Semantic coherence grouping & 68.8 & 42.8 & 62.7 \\
\hline
\end{tabular}
\end{table}

Table~\ref{tab:ablation_results} illustrates the modality effectiveness study. The single-modal baselines establish the foundation for understanding each modality's contribution. Video-only demonstrates stronger cross-subject performance compared to other metrics, but suffers significantly in cross-view scenarios, particularly CV1, highlighting view variance challenges. Pose-only shows more consistent performance across protocols due to its inherent view-invariant properties.

The bi-modal combinations reveal complementary relationships between modalities. Video + Pose achieves the most significant improvement over individual modalities, demonstrating that geometric information effectively complements visual features for person-independent recognition. Video + Object provides moderate improvements, with object context helping disambiguate activities with similar motion patterns.

The full multi-modal system achieves optimal performance across all protocols, validating the fusion architecture effectiveness. 
Removing pose normalization results in notable performance degradation, particularly in cross-view scenarios, confirming the critical importance of view-invariant preprocessing for robust indoor activity recognition. This analysis confirms that each component serves a distinct and essential role in the multi-modal framework, with the full system achieving a balanced integration of all three modalities for optimal AAL activity recognition performance.

To assess the effectiveness of our grouping strategy, we compared it against a baseline that forms groups without considering cross-activity co-occurrence (i.e., ignoring how often objects appear together in the same videos). Our few-coincidences approach, merging objects that rarely co-occur across activities, achieved higher accuracy while operating under the same computational budget (eight masks). By minimising within-group co-activation, it reduces redundancy and yields more discriminative group-level representations, better capturing the variety of object–activity interactions in daily living.

We also study the effect of the number of attention heads in the cross-attention module (Table~\ref{tab:attention_heads}), in line with standard Transformer practice. Increasing heads from 2 to 8 yields consistent gains, with the best average performance at 8 heads. Using 16 heads offers no consistent improvement, while incurring additional compute. We therefore adopt 8 heads as a balanced choice.

\begin{table}[t]
\centering
\caption{Performance comparison across different numbers of attention heads showing accuracy results on CS, CV1, and CV2 datasets.}
\label{tab:attention_heads}
\begin{tabular}{p{8cm}@{\hspace{.8cm}}c@{\hspace{.8cm}}c@{\hspace{.8cm}}c}
\textbf{Number of Heads} & \textbf{CS} & \textbf{CV1} & \textbf{CV2}  \\
\hline
2 heads  & 69.2 & 43.5 & 64.8  \\
4 heads  & 69.4 & 43.2 & 65.1    \\
8 heads  & 70.1 & 44.2 & 65.4 \\
16 heads & 69.8 & 43.9 & 65.6  \\
\hline
\end{tabular}
\end{table}

\begin{figure}[t]
\centering
\includegraphics[width=\textwidth]{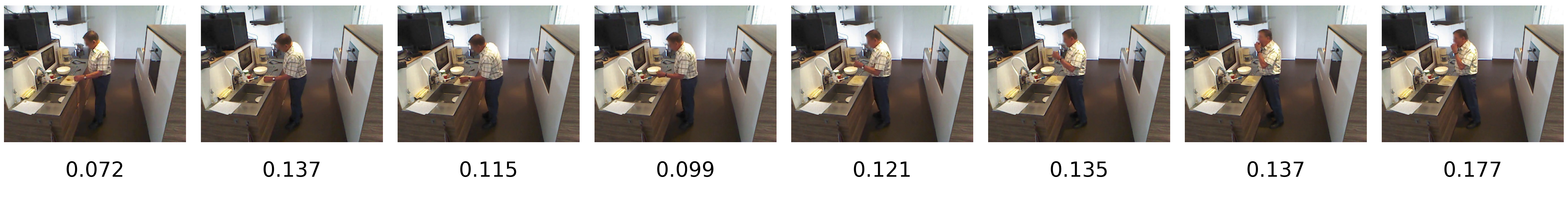}
\caption{Sample frames from the intervals corresponding to temporal attention weights $\alpha_t$ values for activity class "Eating Snacks".}
\label{fig:sample_frames_attention}
\end{figure}

Figure~\ref{fig:sample_frames_attention} illustrates the learned temporal attention weights alongside corresponding video frames, demonstrating how our pose-driven temporal attention mechanism focuses on activity-relevant temporal segments. The visualization shows that the attention mechanism successfully identifies key motion phases, such as the preparation, execution, and completion stages of activities like "Eating Snacks," confirming the effectiveness of our pose-guided temporal weighting strategy.

\section{Conclusion}

This paper presents an effective multi-modal approach for recognizing daily activities in indoor assisted living environments. By integrating video data, human pose information, and object context through a carefully designed cross-attention mechanism, the system achieves robust and accurate classification despite challenges such as viewpoint variance, inter-class similarity, and complex indoor scenes. The model benefits from the complementary strengths of each modality, allowing it to capture rich semantic and temporal cues critical for fine-grained activity understanding. Experimental results on the benchmark dataset demonstrate consistent improvements over single- and multi-modal baselines as well as recent transformer-based approaches, highlighting the value of leveraging multi-modal fusion with spatial-temporal attention.

Looking forward, an important research direction is to reduce the dependence on multiple modalities during inference, thereby simplifying deployment and improving efficiency. Developing modality-agnostic representations or knowledge distillation methods that allow the system to operate effectively using only RGB input while retaining performance is a promising avenue. In addition, exploring self-supervised or unsupervised learning techniques could substantially lessen reliance on annotated data and improve the model’s adaptability to new environments and subjects. Further improvements might come from designing adaptive attention mechanisms that dynamically allocate computational resources based on scene complexity or activity context, balancing performance and efficiency. Investigating domain adaptation and lifelong learning strategies would also advance the system's generalization to diverse real-world conditions.

Moreover, future work on the object modality could focus on more efficient grouping and attention mechanisms to better capture the spatial and temporal relationships between humans and relevant objects without significantly increasing computational costs. With respect to transformer architectures, while recent advances show promise, challenges remain concerning their high computational demand and data requirements. Developing hybrid models that integrate transformer-based attention within lighter multi-modal frameworks, combined with pretraining strategies, may yield more practical and effective systems for assisted living scenarios.

By addressing these directions, future work can contribute to more scalable, privacy-aware, and intelligent activity recognition solutions that meet the needs of real-world assisted living scenarios.

\bmhead{Acknowledgements}

This work is part of the visuAAL project on Privacy-Aware and Acceptable Video-Based Technologies and Services for Active and Assisted Living~(\url{https://www.visuaal-itn.eu/}). This project has received funding from the European Union’s Horizon 2020 research and innovation programme under the Marie Skłodowska-Curie grant agreement No. 861091.


\bibliography{sn-bibliography}

\end{document}